\documentclass[sigconf]{acmart}
\usepackage{hyperref}
\usepackage{algorithm}
\usepackage{algorithmic}
\usepackage{booktabs}
\usepackage{bm}
\AtBeginDocument{%
  \providecommand\BibTeX{{%
    \normalfont B\kern-0.5em{\scshape i\kern-0.25em b}\kern-0.8em\TeX}}}

\setcopyright{acmcopyright}
\copyrightyear{2018}
\acmYear{2018}
\acmDOI{10.1145/1122445.1122456}

\acmConference[Woodstock '18]{Woodstock '18: ACM Symposium on Neural
  Gaze Detection}{June 03--05, 2018}{Woodstock, NY}
\acmBooktitle{Woodstock '18: ACM Symposium on Neural Gaze Detection,
  June 03--05, 2018, Woodstock, NY}
\acmPrice{15.00}
\acmISBN{978-1-4503-XXXX-X/18/06}

\begin{document}

\title{Discovering Dialog Structure Graph for Open-Domain Dialog Generation}

%
% The "author" command and its associated commands are used to define
% the authors and their affiliations.
% Of note is the shared affiliation of the first two authors, and the
% "authornote" and "authornotemark" commands
% used to denote shared contribution to the research.

% \affiliation{%

% \author{
% 	Jun Xu,\textsuperscript{\rm 1}
% 	Haifeng Wang,\textsuperscript{\rm 2}
% 	Zhengyu Niu,\textsuperscript{\rm 2}
% 	Hua Wu,\textsuperscript{\rm 2}
% 	Wanxiang Che\textsuperscript{\rm 1}\thanks{Corresponding author}\\ % All authors must be in the same font size and format. Use \Large and \textbf to achieve this result when breaking a line
% \textsuperscript{\rm 1}Harbin Institute of Technology, Harbin, China\\
% \textsuperscript{\rm 2}Baidu Inc., Beijing, China\\ %If you have multiple authors and multiple 
% \{jxu, car\}@ir.hit.edu.cn, \{wanghaifeng, niuzhengyu, wu\_hua\}@baidu.com % email address must be in roman text type, not monospace or sans serif
% }

\author{Jun Xu}
\authornote{This work was done at Baidu.}
\email{jxu@ir.hit.edu.cn}
\affiliation{%
  \institution{Harbin Institute of Technology}
  \country{China}
}

\author{Zeyang Lei,Haifeng Wang, Zheng-Yu Niu,
        \\Hua Wu}
\email{{wanghaifeng, niuzhengyu, wu\_hua, huangjizhou01}@baidu.com}
\affiliation{%
  \institution{Baidu, Inc.}
  \country{China}
}

\author{Wanxiang Che}
\authornote{Corresponding author.}
\email{car@ir.hit.edu.cn}
\affiliation{%
  \institution{Harbin Institute of Technology}
  \country{China}
}

\author{Ting Liu}
\email{tliu@ir.hit.edu.cn}
\affiliation{%
  \institution{Harbin Institute of Technology}
  \country{China}
}

% \affiliation{%

% \affiliation{%

% \affiliation{%

% \affiliation{%

% \affiliation{%

\renewcommand{\shortauthors}{Trovato and Tobin, et al.}

\begin{abstract}
	Learning interpretable dialog structure from human-human dialogs yields basic insights into the structure of conversation, and also provides background knowledge to facilitate dialog generation. In this paper, we conduct unsupervised discovery of dialog structure from chitchat corpora, and then leverage it to facilitate dialog generation in downstream systems. To this end, we present a Discrete Variational Auto-Encoder with Graph Neural Network (DVAE-GNN), to discover a unified human-readable dialog structure. The structure is a two-layer directed graph that contains session-level semantics in the upper-layer vertices, utterance-level semantics in the lower-layer vertices, and edges among these semantic vertices. In particular, we integrate GNN into DVAE to fine-tune utterance-level semantics for more effective recognition of session-level semantic vertex. Furthermore, to alleviate the difficulty of discovering a large number of utterance-level semantics, we design a coupling mechanism that binds each utterance-level semantic vertex with a distinct phrase to provide prior semantics. Experimental results on two benchmark corpora confirm that DVAE-GNN can discover meaningful dialog structure, and the use of dialog structure graph as background knowledge can facilitate a graph grounded conversational system to conduct coherent multi-turn dialog generation. 
\end{abstract}




\maketitle

\section{Introduction}
With the aim of building a machine to converse with humans naturally, some work investigate neural models \cite{serban2017hierarchical,Shang2015} to build chatbots with large-scale human-human dialog corpora that are collected from various websites, especially those social media websites. While these models can generate locally relevant dialogs, they struggle to organize individual utterances into globally coherent ﬂow. The possible reason is that it is difﬁcult to control the overall dialog ﬂow without background knowledge about dialog structure. However, due to the complexity of open-domain conversation, it is impossible to design dialog structure manually. Therefore, it is of great importance to automatically discover interpretable open-domain dialog structure from dialog corpus for dialog generation.

%\cite{hirano2016analyzing}

To the best of our knowledge, there is no previous work proposed to discover open-domain dialog structure. The most related work are previous studies that tried to discover dialog structure from task-oriented dialogs~\cite{Shi2019}.
%The most related work is Shi \textit{et al.} \cite{Shi2019}, which tried to discover dialog structure from task-oriented dialogs.
However, they faced \textbf{two limitations} when applied to discover open-domain dialog structure.  \textbf{First}, it only discovers utterance-level structure, where each vertex represents fine-grained semantic at the level of utterance. 
%This is reasonable in task-oriented dialogs as they share the same session-level semantic (or topics), which is the task itself. 
However, besides utterance-level semantics, session-level goals (chatting topics) are essential for open-domain dialog generation~\cite{Wu2019,Kang2019,Moon2019}.\footnote{Here goals refer to, what participants expect to accomplish throughout a dialog session, e.g., greeting or invitation.} To provide a full picture of open-domain dialog structure, it is desirable to provide a hierarchical dialog structure that represents semantics (and their transitions) at the two levels of session and utterance. 
\textbf{Second}, the method is designed for task-oriented dialogs, and the discovered dialog structure contains only a few or a dozen discrete semantics. It limits the ability to discover large-scale (more than one million in this paper) semantics that exist in open-domain dialogs.

\begin{figure*}[htbp]
	\centering
	\includegraphics[width = \textwidth]{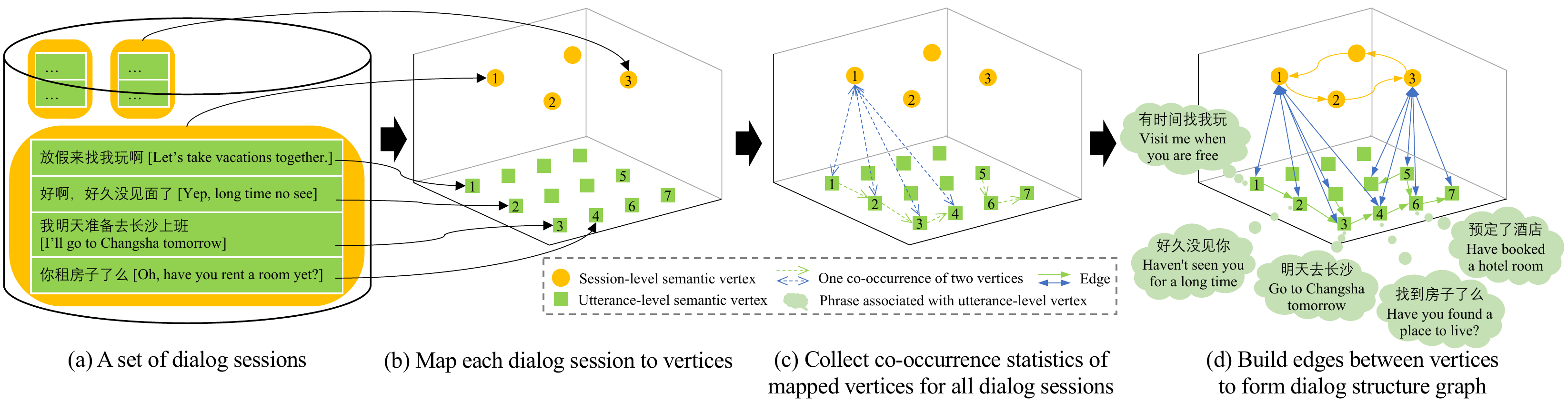}
	\caption{The procedure of dialog structure discovery. Figure (d) shows the discovered \textbf{unified} dialog structure graph.}
	\label{fig:procedure}
\end{figure*}

In this paper, we try to discover a \textbf{unified} human-readable dialog structure from all dialog sessions in the corpus.
The structure is a two-layer directed graph that contains session-level semantics in the upper-layer vertices, utterance-level semantics in the lower-layer vertices, and edges among these semantic vertices. 
To this end, we propose a novel Discrete Variational Auto-Encoder with Graph Neural Network (\textbf{DVAE-GNN}).

To address the \textbf{first limitation} mentioned above, we design the DVAE-GNN  model to discover a dialog structure that captures both session-level and utterance-level semantics. It contains a recognition procedure that maps a dialog session to session-level and utterance-level semantic vertices, and a reconstruction procedure that regenerates the entire dialog session with these mapped semantic vertices, as similar to \cite{van2017neural}. Specifically, given a dialog session, DVAE-GNN maps each utterance in the session to an utterance-level semantic vertex at the lower layer, and then maps the entire session to a session-level semantic vertex at the upper layer with the help of GNN.

\begin{figure}[htbp]
	\centering
	\includegraphics[width = 0.4\textwidth]{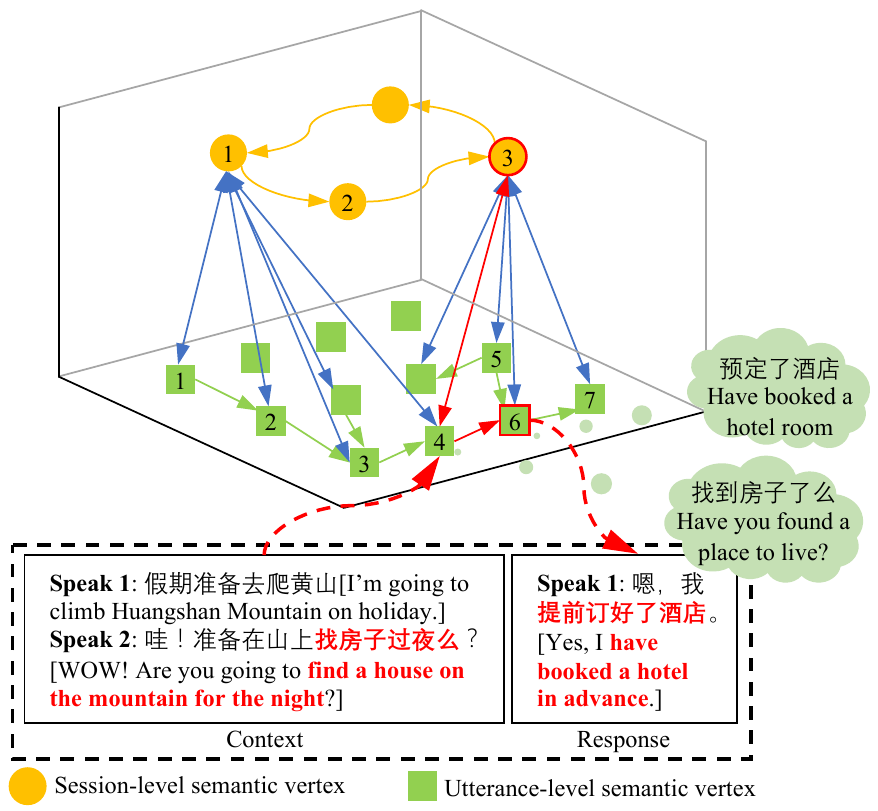}
	\caption{Dialog structure grounded response generation.}
	\label{fig2}
\end{figure}

To address the \textbf{second limitation} mentioned above, we devise a coupling mechanism to provide prior semantic information for each utterance-level semantic vertex, through binding each utterance-level semantic vertex with a distinct phrase extracted from corpus. Specifically, the representation of each utterance-level semantic vertex in the graph is composed of the representation vector of one distinct phrase and another learnable latent vector. 
Intuitively, with such phrase, there is no need to learn utterance-level semantic representation from the scratch. Thus, the learning difficulty is reduced, which makes it much easier for discovering large-scale (more than one million) utterance-level semantics in corpus.

As shown in Figure \ref{fig:procedure}, we present how to discover the two-layer dialog structure graph. Given a set of dialog sessions in Figure \ref{fig:procedure} (a), we map each dialog session to semantic vertices at the two levels, by running the recognition procedure of DVAE-GNN, as shown in Figure \ref{fig:procedure} (b). Then we collect co-occurrence statistics of mapped vertices in dialog sessions. As shown in Figure \ref{fig:procedure} (c), each green directional arrow denotes one co-occurrence of two utterance-level semantic vertices, which represents two utterances' ordering relation. And each blue bidirectional arrow denotes one co-occurrence of an utterance-level semantic vertex and a session-level semantic vertex, which represents the utterance's membership in the session.
Finally, we build edges among vertices based on all collected co-occurrence statistics to form the unified dialog structure graph, as shown in Figure \ref{fig:procedure} (d).

\iffalse
In Figure \ref{fig:procedure}-(c), each arrow denotes once co-occurrence of two vertices, which represents two utterances' ordering relation (a black directional arrow) or an utterance's membership in a session (a blue bidirectional arrow). 
\fi

To leverage this graph for conversation generation, we propose a graph grounded conversational system (\textbf{GCS}). As shown in Figure \ref{fig2}, given a dialog context (the last two utterances), GCS first maps it to utterance-level semantic vertex, and then learns to walk over graph edges, and finally selects a contextual appropriate utterance-level semantic vertex to guide response generation at each turn.

Our contribution is summarized as follows:
\begin{itemize}
	\item We identify the task of discovering a two-layer dialog structure graph from open-domain dialogs. 
	\item We propose an unsupervised DVAE-GNN model for open-domain dialog structure discovery. 
	Its novelty lies in two strategies: (1) we integrate GNN into VAE to fine-tune utterance-level semantics for more effective recognition of session-level semantic vertex, and (2) we design a coupling mechanism that binds each utterance-level vertex with a distinct phrase to help solve the severe learning difficulty when discovering a large number of semantics in VAE.
	\item Results on two benchmark dialog datasets demonstrate that (1) DVAE-GNN can discover meaningful dialog structure, (2) the use of GNN and phrases are crucial to the success of dialog structure discovery, and (3) the dialog structure graph can enhance dialog coherence.
\end{itemize}

\section{Related Work}

\begin{figure*}[htbp]
	\centering
	\includegraphics[width = \textwidth]{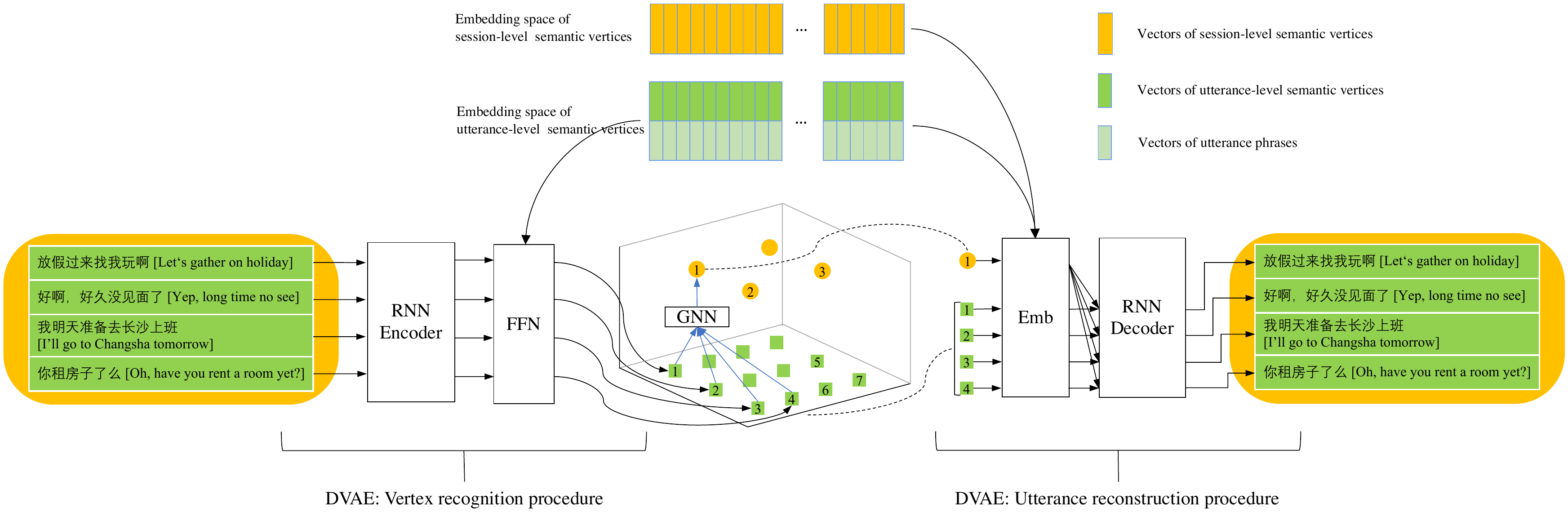}
	\caption{Overview of DVAE-GNN. FFN denotes feed-forword neural networks and Emb refers to embedding layers}
	\label{fig:DVAE-overview}
\end{figure*}

\subsection{Dialog Structure Learning for Task-Oriented dialogs} 
There are previous work on discovering human-readable dialog structure for task-oriented dialogs \cite{Chotimongkol2008,Zhai2014,Tang2018,Shi2019}. Specifically, the hidden Markov model is used to capture the temporal dependency within human dialogs~\cite{Chotimongkol2008}. Furthermore, Tang et al.~\cite{Tang2018} propose to learn sub-goals of task dialogs from a set of state trajectories generated by another rule-based agent, instead of from dialog utterances, which limits their application on open-domain dialogs. Moreover, Shi et al.~\cite{Shi2019} proposes to  extract dialog structures using a modified variational recurrent neural network (VRNN) model with discrete latent vectors. 

However, the number of semantics (vertices) involved in the dialog structures discovered by these models are limited to only dozens or hundreds, which cannot cover fine-grained semantics in open-domain dialogs. In this paper, we propose a novel DVAE-GNN model to discover a large-scale dialog structure graph (more than one million vertices) from open-domain dialog corpus. Furthermore, our method can discover a hierarchical dialog structure, which is different from the non-hierarchical dialog structures in previous work. 

\iffalse
\begin{algorithm} 
%\small
 \caption{Phrase extraction}
  \label{event-extract}
 \begin{algorithmic}[1]
 \REQUIRE 
     An utterance $U$
 \ENSURE 
     A set of phrases $E$ extracted from $U$
 \STATE 
Obtain a dependency parse tree $T$ for $U$;
 \STATE Get all the head words {HED} that are connected to {ROOT} node, and all the leaf nodes in $T$ (denoted as $L$);
 \FOR{each leaf node in $|L|$}
 \STATE Extract a phrase consisting of words along the tree from {HED} to current leaf node, denoted as $e_i$;
 \STATE If $e_i$ is a verb phrase, then append it into $E$;
 \ENDFOR
 \RETURN $E$
 \end{algorithmic}
\end{algorithm}
\fi

\subsection{Open-Domain Dialog Systems}
Early work on conversational systems employ hand-crafted rules or templates to build chatbots \cite{1,wallace2009anatomy}. Recently, some work propose retrieval-based models \cite{ji2014information,yan2016learning,wu2017sequential}, generation-based models \cite{Serban2016,li2016deep,13,Zhou2017Emotional,Mazumder2018Towards,shao2017generating,ACL,chen2019generating,Tian2019} or the fusion of the two lines \cite{8,zhuang2017ensemble} to construct neural dialog systems. 

To encourage diverse responses, some studies have been conducted to improve response informativeness by integrating additional keywords ~\cite{yao2017towards,xing2017topic,Chat2018Wang} or latent variables \cite{gao2019discrete,Ghandeharioun2019,Park2018,shenetal2019,serban2017hierarchical,zhao2017learning,zhao2018unsupervised,xu-etal-2020-conversational} into the process of response generation. Specifically, previous work studies the one-to-many mapping problem for dialog modeling, by introducing either a Gaussian latent distribution with Conditional Variational Auto-Encoder (CVAE) \cite{serban2017hierarchical,zhao2017learning} or multiple latent mechanisms \cite{tao2018get,zhou2017mechanism,chen2019generating}.

In our work, we focus on hierarchical dialog structure modeling, not limited to utterance-level modeling in previous work. We notice that a variational hierarchical RNN based conversation model is proposed in \cite{Park2018} to learn both local and global latent variables from dialogs. Our work differs from theirs in that: (1) we aim at discovering dialog structure from dialogs, while they try to address the degeneration problem in variational RNNs; (2) our vertices are discrete and the majority of them are (utterance-level semantic vertices) interpretable, and we organize them as a graph to facilitate downstream systems. Their latent variables are continuous, difﬁcult to interpret, and it is hard to build a graph with their variables.

\section{Our Approach}
We first define the task of open-domain dialog structure discovery in Section \ref{problem definition}. Then, we describe the process of dialog structure discovery, including the DVAE-GNN model (from Section \ref{sec::initialize} to Section \ref{sec::loss}) and graph construction with well-trained DVAE-GNN (in Section \ref{sec:graphconstruction}). Finally, we propose a graph grounded conversational system (GCS) to utilize the discovered dialog structure graph to enhance dialog coherence in Section \ref{gcs}.

Specifically, the proposed DVAE-GNN model contains two procedures: (1) a vertex recognition procedure that first maps each utterance in a dialog session to an utterance-level semantic vertex in the graph, and then maps the entire dialog session to a session-level semantic vertex in the graph; (2) an utterance reconstruction procedure that regenerates all utterances in the dialog session with the mapped utterance-level and session-level semantic vertices. Here, the utterance reconstruction procedure is utilized to train the recognition stage. Intuitively, in order to reconstruct all utterances in the dialog session as much as possible, DVAE-GNN is required to (1) discover representative semantics at both levels accurately, and (2) map each utterance in the session and the entire session itself to semantic vertices in the graph exactly.

\subsection{Task Definition}
\label{problem definition}
Given a corpus $D$ that contains $|D|$ dialog sessions $\{X_1, X_2, ..., X_{|D|}\}$, where each dialog session $X$ consists of a sequence of $c$ utterances, and $X=[x_1, ..., x_c]$. The objective is to discover a \textbf{unified} two-layer dialog structure graph $\mathcal{G}=\{\mathcal{V},\mathcal{E}\}$ from all dialog sessions in $D$. Here, $\mathcal{V}$ is the vertex set and $\mathcal{E}$ is the edge set. 

As shown in Figure \ref{fig:procedure} (d), the graph consists of two types of vertices and three types of edges. The details are as follows.

(1) {Vertices}: There are two types of vertices, session-level semantic vertices and utterance-level semantic vertices. Let $v^s_m$ ($1\leq m\leq M$) denote session-level semantic vertices (goals) and let $v^u_n$ ($1\leq n\leq N$) denote utterance-level semantic vertices. $M$ and $N$ refer to the number of two types of vertices respectively. 

\begin{algorithm} 
%\small
 \caption{Phrase extraction}
  \label{event-extract}
 \begin{algorithmic}[1]
 \REQUIRE 
     An utterance $U$
 \ENSURE 
     A set of phrases $E$ extracted from $U$
 \STATE 
Obtain a dependency parse tree $T$ for $U$;
 \STATE Get all the head words HED that are connected to ROOT node, and all the leaf nodes in $T$ (denoted as $L$);
 \FOR{each leaf node in $|L|$}
 \STATE Extract a phrase consisting of words along the tree from HED to current leaf node, denoted as $e_i$;
 \STATE If $e_i$ is a verb phrase, then append it into $E$;
 \ENDFOR
 \RETURN $E$
 \end{algorithmic}
\end{algorithm}

(2) {Edges}: Three types of edges contain edges between two session-level semantic vertices (denoted as Sess-Sess edges), edges between two utterance-level semantic vertices (denoted as Utter-Utter edges), and edges between an utterance-level semantic vertex and its parent session-level semantic vertices (denoted as Sess-Utter edges).
Here, Sess-Sess edges or Utter-Utter edges indicate natural dialog transitions at different levels, i.e. session-level goal transition or utterance-level dialog content transition. The Sess-Utter edges indicate the utterance-level semantic vertices are fine-grained semantic information about their parent session-level semantic vertices (goals). Here it is allowed that each utterance-level semantic vertex can have multiple session-level semantic vertices as parents.

\subsection{Initialization}
\label{sec::initialize}
To make it easier to learn dialog structure graph, we initialize the dialog structure graph with a coupling mechanism that binds each utterance-level vertex with a distinct phrase.

\textbf{{Vertex Initialization.}} We initialize the representation of each session-level vertex with a learnable latent vector. Furthermore, to initialize the representation of each utterance-level semantic vertex, we concatenate two vectors: a learnable latent vector and a distinct phrase's representation vector. Here, a distinct phrase that coupled with an utterance-level semantic vertex is denoted as the utterance-level semantic vertex's associated phrase. Specifically, these associated phrases are obtained by first extracting phrases from corpus with Algorithm \ref{event-extract}, and then binding the $n$-th utterance-level semantic vertex with the $n$-th most frequent phrase. Intuitively, these associated phrases provide prior semantic knowledge for embedding space learning of utterance-level semantic vertices, which helps train the DVAE-GNN model effectively.

Formally, we use $\bm{\Lambda}_g$ to represent the embedding matrix of session-level semantic vertices and $\bm{\Lambda}_x$ to represent the embedding matrix of utterance-level semantic vertices. The representation vector $\bm{\Lambda}_g[m]$ of $m$-th session-level semantic vertex is one learnable latent vector. 
The representation vector of $n$-th utterance-level semantic vertex can be calculated as follows.
\begin{equation}
\bm{\Lambda}_x[n] = W^{u}[\bm{e}(ph_n);\bm{v}_n]
\end{equation}
where $\bm{v}_n$ is a learnable latent vector, $\bm{e}(ph_n)$ denotes the representation vector of the associated phrase $ph_n$, and ``$;$'' denotes concatenation operation. For phrase representation, 
we first feed words in associated phrase to an RNN and obtain their hidden states. Here, the RNN is denoted as the phrase encoder. Then we compute the average pooling value of these hidden states as $\bm{e}(ph_n)$.
Specifically, parameters in all learnable latent vectors are randomly initialized.

\textbf{{Edge Initialization}} We build initial Utter-Utter edge between two utterance-level semantic vertices when their associated phrases can be extracted sequentially from two adjacent utterances in the same session. The other two types of edges (Sess-Sess and Sess-Utter edges) will be dynamically learned by the DVAE-GNN model in the following section. 

Here, we denote a graph that consists of only phrases (as vertices) and these initial edges (between phrases) as \textbf{Phrase Graph}.
\iffalse
We first initialize each of the $M$ upper-layer vertices and $N$ lower-layer vertices with a learnable vector. Then, we extract all phrases from corpus with Algorithm \ref{event-extract}, and then only keep the top-$N$ most frequent phrases. Furthermore, we bind each lower-layer vertex with a distinct phrase to provide prior semantic. Last, we build an initial direct edge between two lower-layer vertices if their bound phrases can be extracted sequentially from two adjacent utterances in corpus for more than 3 times. 

Here, we denote a graph that consists of only phrases (vertices) and these initial edges (between phrases) as \textbf{Phrase Graph}.
\fi

\subsection{Vertex Recognition}
\textbf{Utterance-level Vertex Recognition.}
\label{sec:rec_u_state}
For each utterance $x_i$ in a dialog session, we map it to an utterance-level semantic vertex. To be specific, we first encode the utterance $x_i$ with an RNN encoder to obtain its representation vector $\bm{e}(x_i)$. 
Then, we calculate the utterance-level posterior distribution by a feed-forward neural network (FFN):
\begin{equation}
    \label{inf_utter}
    z_i \sim q(z|x_i) = \text{softmax}(\bm{\Lambda}_x\bm{e}(x_i)).
\end{equation} 
Finally, we obtain the mapped utterance-level semantic vertex, $z_i$, by sampling from the posterior distribution with gumbel-softmax~\cite{JangCategorical}. Here, we can obtain an utterance-level semantic vertex sequence after mapping each utterance in one dialog session, where the utterance-level semantic vertex sequence is utilized for session-level semantic vertex recognition.

In this work, it is costly to calculate Equation \ref{inf_utter} since the total number of utterance-level semantic vertices, $N$, is very large (more than one million). In practice, for each utterance, we first retrieve the top-50 most related utterance-level semantic vertices according to Okapi BM25~\cite{robertson2009probabilistic} similarity between the utterance and associated phrases of all candidate vertices. And then calculate Equation \ref{inf_utter} only with these retrieved vertices. 
Thus, only a part of vectors in $\bm{\Lambda}_x$ will be dynamically updated for each training sample when training DVAE-GNN.

\noindent\textbf{Session-level Vertex Recognition.}
\label{sec:rec_s_state}
We assume that each session-level semantic vertex (dialog goal) corresponds to a group of similar utterance-level semantic vertex sequences that are mapped by different dialog sessions. And these similar vertex sequences might have overlapped utterance-level semantic vertices. To encourage mapping similar utterance-level semantic vertex sequences to similar session-level semantic vertices, we employ GNN to propagate local information to the whole sequences. Specifically, we utilize a three-layer graph convolution network (GCN) over Utter-Utter edges to calculate structure-aware utterance-level semantics. The calculation is defined by:
\begin{equation}
    \bm{h}^{j}_{v^u_n} = \sigma^j(\sum_{v^{u}_{n'} \in  \mathcal{N}(v^u_n)} \bm{h}^{j-1}_{v^{u}_{n'}}),
\end{equation}
where $\bm{h}^{j}_{v^u_n}$ denotes the $j$-th layer structure-aware representation for the $n$-th utterance-level semantic vertex $v^{u}_n$. $\sigma^{j}$ is the $sigmoid$ activation function for the $j$-th layer, and $\mathcal{N}(v^{u}_n)$ is the set of utterance-level neighbors of $v^{u}_n$ in the graph. Here, we can obtain a structure-aware semantic sequence [$\bm{h}^{3}_{v^u_{z_1}}, \bm{h}^{3}_{v^u_{z_i}}, ..., \bm{h}^{3}_{v^u_{z_c}}$], where $\bm{h}^{3}_{v^u_{z_i}}$ represents the final structure-aware representation of $i$-th mapped utterance-level semantic vertex, $v^u_{z_i}$.

Then, we feed the structure-aware semantic sequence to an RNN encoder, denoted as the vertex-sequence encoder, to obtain the structure-aware session representation $\bm{e}(z_{1,...,c})$. Furthermore, we calculate the session-level posterior distribution as follows:
\begin{equation}
    g \sim q(g|z_{1,...,c}) = \text{softmax}(\bm{\Lambda}_g\bm{e}(z_{1,...,c})).
\end{equation}
Moreover, we obtain the mapped session-level semantic vertex, $g$, by sampling from the session-level posterior distribution with gumbel-softmax~\cite{JangCategorical}.

\subsection{Utterance Reconstruction}
\label{sec:utt_rec}
We reconstruct all utterances in the dialog session by feeding these mapped utterance-level and session-level semantic vertices into an RNN decoder (denoted as the reconstruction decoder). Specifically, to regenerate utterance $x_i$, we concatenate the vector of mapped utterance-level semantic vertex $\bm{\Lambda}_x[z_i]$ and the vector of the mapped session-level semantic vertex $\bm{\Lambda}_g[g]$, as the initial hidden state of the reconstruction decoder.
%utterance representation $\mathbf{e}(x_i)$, 

\subsection{Loss Function}
\label{sec::loss}
We optimize the proposed DVAE-GNN model by maximizing the classical variational lower-bound \cite{kingma2013auto}. Please refer to Appendix~\ref{sup:deri} for more details.

\iffalse
To better utilize GNN for session-level semantic vertex recognition, we rebuild the lower-layer edges at the end of each epoch. Specifically, we build an directed edge between two lower-layer vertices if they are sequentially mapped by two adjacent utterances in corpus for more than 3 times. 
\fi

\subsection{Graph Construction}
\label{sec:graphconstruction}

After training, the DVAE-GNN model is able to accurately discover representative semantics at different levels and exactly map each utterance in a session and the entire session itself to utterance-level and session-level semantic vertices in the graph. Then, we utilize the well-trained DVAE-GNN to reveal these representative semantics' relations that exist in dialog corpus. Specifically, we map dialog sessions to semantic vertices in the graph, and build edges between semantic vertices based on co-occurrence statistics of mapped vertices. Specifically, as shown in Figure \ref{fig:procedure}, we construct edges with the following steps. 

First, we map all dialog sessions in corpus to vertices, where each utterance in a dialog session is mapped to an utterance-level semantic vertex and the dialog session itself is mapped to a session-level semantic vertex. 

Then, we collect co-occurrence statistics of these mapped vertices. Specifically, we count the total mapped times for each session-level semantic vertex, denoted as $\#(v^s_i)$, and those for each utterance-level semantic vertex, denoted as $\#(v^u_j)$. Furthermore, we collect the co-occurrence frequency of a session-level semantic vertex and an utterance-level semantic vertex that are mapped by a dialog session and an utterance in it respectively, denoted as $\#(v^s_i, v^u_j)$. Moreover, we collect the co-occurrence frequency of two utterance-level semantic vertices that are sequentially mapped by two adjacent utterances in a dialog session, denoted as $\#(v^u_j, v^u_k)$.

Finally, we build edges between vertices based on these co-occurrence statistics. We first build a directed Utter-Utter edge from $v^u_j$ to $v^u_k$ if the bi-gram transition probability $\#(v^u_j, v^u_k)/\#(v^u_j)$ is above a threshold $\alpha^{uu}$. Then, we build a bidirectional Sess-Utter edge between $v^u_j$ 
and $v^s_k$ if the probability $\#(v^s_i, v^u_j)/\#(v^u_j)$ is above a threshold $\alpha^{su}$. 
Moreover, we build a directed Sess-Sess edge from $v^s_i$ to $v^s_o$ if $\#(v^s_i, v^s_o)/\#(v^s_i)$ is above a threshold $\alpha^{ss}$, where the first item $\#(v^s_i, v^s_o)$ is the number of utterance-level semantic vertices that are connected to both session-level semantic vertices. Here, Sess-Sess edges are dependent on Sess-Utter edges.

\subsection{Graph Grounded Conversational System}
\label{gcs}
To use the constructed dialog structure graph for coherent dialog generation, we propose a graph grounded conversation system (GCS). Specifically, we formulate multi-turn dialog generation as a graph grounded Reinforcement Learning (RL) problem, where vertices in the graph serve as RL actions. 

GCS contains three modules: a dialog context understanding module that maps given dialog context to an utterance-level semantic vertex in the graph, a policy that learns to walk over edges to select an utterance-level semantic vertex to serve as response content, and a response generator that generate an appropriate response based on the selected utterance-level semantic vertex. Specifically, to make the selection easier, the policy contains two sub-policies to select an utterance-level semantic vertex within two steps. First, a session-level sub-policy selects a session-level semantic vertex as current dialog goal. Then, an utterance-level sub-policy selects an utterance-level semantic vertex from current dialog goal's child utterance-level semantic vertices.

In the following, we will elaborate the details of GCS.

\subsubsection{\textbf{Dialog Context Understanding}}
\label{state and action}
Given a dialog context (the last two utterances), we first map it to the graph by recognizing the most related utterance-level semantic vertex with the well-trained DVAE-GNN. Here, the recognized utterance-level semantic vertex is denoted as the hit utterance-level semantic vertex.

For policy learning, we build current RL state $s_{l}$ at time step $l$ by collecting dialog context (the last two utterances), previously selected session-level semantic vertex sequence, and previously selected utterance-level semantic vertex sequence. Here, we first utilize three independent RNN encoders to encode them respectively, and then concatenate these three obtained representation vectors, to obtain the representation of the RL state, $\bm{e}_{s_{l}}$.

\subsubsection{\textbf{Dialog Policy Learning}} \label{D}
For RL based policy training, we utilize a user simulator (i.e., a generative model) to chat with our system GCS. Given a dialog context at each turn, the two sub-policies work together to select an utterance-level semantic vertex for guiding response generation. Then with a response from GCS as input, the user simulator will generate another appropriate response. The dialog between GCS and the user simulator will continue till it reaches the maximum number of turns (8 in this paper), i.e. both of GCS and the user simulator generate 8 utterances.

\textbf{Session-level sub-policy} \label{reward-factor1}
The session-level sub-policy determines the current dialog goal by selecting one of the the hit utterance-level semantic vertex's parent session-level semantic vertices.

Let $\mathcal{A}^{g}_{s_l}$ denote the set of session-level candidate actions at time step $l$. It consists of all parent session-level semantic vertices of the aforementioned hit utterance-level semantic vertex. Given current RL state $s_l$ at the time step $l$, the session-level sub-policy $\mu^{g}$ selects an appropriate session-level semantic vertex from $\mathcal{A}^{g}_{s_l}$ as the current dialog goal. Specifically, $\mu^{g}$ is formalized as follows:
\begin{align}
    \mu^{g}(s_l, v^{s}_{c^g_j})=\frac{\exp({\bm{e}_{s_l}}^T \bm{\Lambda}_g[c^g_j])}{\sum^{N_l^{g}}_{k=1}\exp({\bm{e}_{s_l}}^T \bm{\Lambda}_g[c^g_k])} \nonumber.
\end{align}
Here, $\bm{e}_{s_{l}}$ is the aforementioned RL state representation, $c^g_j$ the $j$-th session-level semantic vertex in $\mathcal{A}^{g}_{s_l}$, and $N_l^{g}$ is the number of session-level semantic vertices in $\mathcal{A}^{g}_{s_l}$.
With the distribution calculated by the above equation, we utilize Gumbel-Softmax~\cite{JangCategorical} to sample a session-level semantic vertex from $\mathcal{A}^{g}_{s_l}$ to serve as current dialog goal. 

\textbf{Utterance-level sub-policy} \label{reward-factor2}
The utterance-level sub-policy selects a contextually appropriate utterance-level semantic vertex from current dialog goal's child utterance-level semantic vertices.

Let $\mathcal{A}^{u}_{s_l}$ denote the set of utterance-level candidate actions at time step $l$. It consists of utterance-level semantic vertices that are connected to the vertex of current dialog goal. Given current state $s_{l}$ at the time step $l$, the utterance-level sub-policy $\mu^{u}$ selects an optimal utterance-level semantic vertex from $\mathcal{A}^{u}_{s_l}$. Specifically, $\mu^{u}$ is defined as follows:
\begin{align}
    \mu^{u}(s_{l}, v^{u}_{c^u_j})=\frac{\exp({\bm{e}_{s_{l}}}^T \bm{\Lambda}_x[c^u_j])}{\sum^{N_l^{u}}_{k=1}\exp({\bm{e}_{s_{l}}}^T\bm{\Lambda}_x[c^u_k])} \nonumber.
\end{align}
Here, $\bm{e}_{s_{l}}$ is the aforementioned RL state representation, $c^u_j$ is the $j$-th utterance-level semantic vertex in $\mathcal{A}^{u}_{s_{l}}$, and $N_l^{u}$ is the number of utterance-level candidate vertices in $\mathcal{A}^{u}_{s_l}$.
With the distribution calculated by the above equation, we utilize Gumbel-Softmax~\cite{JangCategorical} to sample an utterance-level semantic vertex from $\mathcal{A}^{u}_{s_{l}}$, to provide response content.

\begin{table*}[ht] 
	\centering
	\resizebox{1.6\columnwidth}{!}{
	\begin{tabular}{l l c c c c c}
	\toprule
		%\hline
		\multicolumn{1}{l|}{ Datasets} & \multicolumn{1}{l|}{Methods} &  \multicolumn{2}{c|}{  Automatic Evaluation} & \multicolumn{3}{c}{ Human Evaluation}  \\ 
		\cline{3-7} \multicolumn{1}{c|}{}& \multicolumn{1}{c|}{} & NLL & BLEU-1/2. & \multicolumn{1}{|c}{S-U Appr.} & U-U Appr. & Intra-Goal Rele.\\
% 		\multicolumn{1}{l}{Weibo Corpus} & & & & \\
		\hline
		\multicolumn{1}{l|}{Weibo} & \multicolumn{1}{l|}{DVRNN} & 29.187 & 0.427/0.322  &\multicolumn{1}{|c}{-} & 0.16 & - \\
		\multicolumn{1}{l|}{} & \multicolumn{1}{l|}{Phrase Graph} & - & -/-  &\multicolumn{1}{|c}{-} & 0.63 & - \\
		\multicolumn{1}{l|}{}& \multicolumn{1}{l|}{DVAE-GNN} & \textbf{20.969} & \textbf{0.588}/\textbf{0.455}   &\multicolumn{1}{|c}{\textbf{0.85}} & \textbf{0.79} & \textbf{1.44} \\
		\multicolumn{1}{l|}{}& \multicolumn{1}{l|}{DVAE-GNN w/o GNN} & 23.364 & 0.560/0.429   &\multicolumn{1}{|c}{0.53} & 0.78 & 1.06\\
		\multicolumn{1}{l|}{}& \multicolumn{1}{l|}{DVAE-GNN w/o phrase} & 24.282 & 0.468/0.355   &\multicolumn{1}{|c}{0.43} & 0.27 & 0.95 \\
		\hline
% 		\multicolumn{1}{l}{Douban Corpus} & & & & \\
% 		\hline
		\multicolumn{1}{l|}{Douban} & \multicolumn{1}{l|}{DVRNN}& 72.744 & 0.124/0.093  &\multicolumn{1}{|c}{-} & 0.14 & - \\
		\multicolumn{1}{l|}{} & \multicolumn{1}{l|}{Phrase Graph} & - & -/-  &\multicolumn{1}{|c}{-} & 0.34 & - \\
		\multicolumn{1}{l|}{}& \multicolumn{1}{l|}{DVAE-GNN}& \textbf{35.975} & \textbf{0.525}/\textbf{0.412}   &\multicolumn{1}{|c}{\textbf{0.60}} & \textbf{0.70} & \textbf{0.93} \\
		\multicolumn{1}{l|}{}& \multicolumn{1}{l|}{DVAE-GNN w/o GNN}& 37.415 & 0.504/0.394   &\multicolumn{1}{|c}{0.38} & 0.54 & 0.48 \\
		\multicolumn{1}{l|}{}& \multicolumn{1}{l|}{DVAE-GNN w/o phrase}& 49.606 & 0.254/0.206   &\multicolumn{1}{|c}{0.28} & 0.19 & 0.27 \\
	\bottomrule
	\end{tabular}
	}
	\caption{Results of dialog structure discovery on Weibo corpus and Douban corpus. As DVRNN learns only utterance-level states, it results in S-U Appr. and Intra-Goal Rele. are not available.}
	\label{tab_DSGWeibo}
\end{table*}

\textbf{RL rewards} To generate a coherent and diverse dialog, we devise a set of reward factors for the two sub-policies.
For the session-level sub-policy, its reward $r^{g}$ is the average rewards from the utterance-level sub-policy during current dialog goal. 
The reward for the utterance-level sub-policy, $r^{u}$, is a weighted sum of the below-mentioned factors. The default values of weights are set as [60, 0.5, -0.5]. \footnote{We optimize these weights by grid search.}

\ \textbf{i) Utterance relevance} We choose the classical multi-turn response selection model, DAM in \cite{zhou2018multi}, to calculate utterance relevance. We expect the generated response is coherent to dialog context.

\ \textbf{ii) Utter-Goal closeness} The selected utterance-level semantic vertex $v^u_j$ should be closely related to current goal $v^s_i$. And we use the $\#(v^s_i, v^u_j)/\#(v^u_j)$ in Section~\ref{sec:graphconstruction} as the utter-goal closeness score.

\ \textbf{iii) Repetition penalty} This factor is 1 when the selected utterance-level semantic vertex shares more than $60\%$ words with one of contextual utterance, otherwise 0. We expect that the selected utterance-level semantic vertices are not only coherent, but also diverse.

More details can be found in the Section \ref{sec:exp_set}.

\subsubsection{Response Generator}
\label{generator}
The response generator is a pre-trained Seq2Seq model with attention mechanism, whose parameters are not updated during RL training. Specifically, we take the last user utterance, and the associated phrase of the selected utterance-level semantic vertex as input of the generator.

\section{Experiments for Dialog Structure Graph Discovery}
\subsection{Datasets and Baselines} 
We evaluate the quality of dialog structure graph discovered by our method and baselines on two benchmark datasets: (1) \textbf{Weibo}~\cite{li2018overview}: this is a Chinese multi-turn tweet-style corpora. After data cleaning, we obtain 3.1 million sessions for training, 10k sessions for validation and 10k sessions for testing.  (2) \textbf{Douban}~\cite{wu2016sequential}: we use the original multi-turn dialog corpus, and obtain 2.3 million sessions for training, 10k sessions for validation and 10k sessions for testing. 
For the Weibo or Douban corpus, each session has 4 sentences on average, and each sentence contains about 7 or 14 words respectively. The discovered dialog structure graph on Weibo corpus contains 1,641,238 utterance-level semantic vertices, 6000 session-level semantic vertices and 11,561,007 edges. And the discovered dialog structure graph on Douban corpus contains 1,768,720 utterance-level semantic vertices, 5500 session-level semantic vertices and 6,117,159 edges. The number of session-level semantic vertices is determined by grid search based on the NLL metric in Section \ref{Evaluation Metrics 1}.

%\subsection{Baseline}
In this work, we select DVRNN \cite{Shi2019}, originally proposed for discovering dialog structure from task dialogs, as a baseline, since there is no previous study on discovering human-readable dialog structure from open-domain dialogs. We rerun the source codes released by the original authors.\footnote{github.com/wyshi/Unsupervised-Structure-Learning} Note that, to suite the setting of our task (open-domain dialog) and also consider the limit of our 16G GPU memory (we set batch size as 32 to ensure training efficiency), we set the number of dialog states as 50 (originally it is 10).\footnote{We ever tried to modify their codes to support the learning of a large number of dialog states (up to 30k). But its performance is even worse than original code with 50 states.} We also evaluate the initial Phrase Graph in Section~\ref{problem definition}.

\subsection{Evaluation Metrics}
\label{Evaluation Metrics 1}
For automatic evaluation, we use two metrics to evaluate the performance of reconstruction: (1) \textbf{NLL} is the negative log likelihood of dialog utterances; (2) \textbf{BLEU-1/2} measures how much that reconstructed sentences contains 1/2-gram overlaps with input sentences~\cite{papineni2002bleu}. The two metrics indicate how well the learned dialog structure graph can capture representative semantics in dialogs. 

Moreover, to evaluate quality of edges in the discovered graph, we conduct human evaluation: (1) \textbf{S-U Appr.} It measures the appropriateness of Sess-Utter edges between one utterance-level semantic vertex and its parent session-level semantic vertices, which is important for dialog coherence (see results in Section \ref{dialog-aba}). ``1'' if an utterance-level vertex is relevant to its session-level vertex (goal), otherwise ``0''. (2) \textbf{U-U Appr.} It measures the quality of Utter-Utter edges between two utterance-level vertices, which is important for response appropriateness of downstream systems. It is ``1'' if an utterance-level edge is suitable for responding, otherwise ``0''. Furthermore, we measure the relevance between a session and its' mapped goal directly. Note that we don't evaluate the quality of Sess-Sess edges because Sess-Sess edges are dependent on the statistics of Sess-Utter edges.

Meanwhile, to evaluate the relevance of sessions about a goal (session-level semantic vertex), we also conduct human evaluation: \textbf{Intra-Goal Rele.} It measures the semantic distance between the two sessions within the same goal label.  
It is ``2'' if two sessions with the same goal label are about the same or highly related goal, ``1'' if two sessions are less relevant, otherwise ``0''. Specifically, during evaluation, we provide typical words of each goal (or utterance-level states in DVRNN) that are calculated based on TF-IDF.
High intra-goal relevance can foster more intra-topic coherent dialogs.
For human evaluation, we sample 300 cases and invite three annotators from a crowd-sourcing platform to evaluate each case.\footnote{test.baidu.com} Notice that all system identifiers are masked during human evaluation. 

\subsection{Implementation Details}
For all models, we share the same vocabulary (maximum size is 50000) and initialized word embedding (dimension is 200)  with the pre-trained Tencent AI Lab Embedding. \footnote{ai.tencent.com/ailab/nlp/embedding.html}
Meanwhile, we randomly initialized the embedding space of session-level semantic vertices and latent vectors for utterance-level semantic vertices (dimensions are 200). The hidden sizes of all RNN encoders and RNN decoders are set as 512. The three threshold variables about co-occurrence statistics $\alpha^{uu}$, $\alpha^{su}$ and $\alpha^{ss}$ are all set as 0.05.

\subsection{Experiment Results}
%\noindent \textbf{Results} 
As shown in Table \ref{tab_DSGWeibo}, DVAE-GNN significantly outperforms DVRNN, in terms of all the metrics (sign test, p-value $<$ 0.01) on the two datasets. It demonstrates that DVAE-GNN can better discover meaningful dialog structure graph that models the interaction logic in dialogs. Specifically, DVAE-GNN obtains the best results in terms of NLL and BLEU-1/2, which demonstrates that DVAE-GNN achieves better reconstruction ability in comparison with DVRNN. Meanwhile, DVAE-GNN also surpasses all baselines in terms of U-U Appr. It indicates that our discovered dialog structure graph has higher quality edges and can better facilitate coherent dialog generation.

Furthermore, we also conduct ablation study. In particular, to evaluate the contribution of GNN, we remove GNN from DVAE-GNN, denoted as DVAE-GNN w/o GNN. We see that its performance in terms of ``S-U Appr.'' and ``Intra-Goal Rele.'' drop sharply. It demonstrates the importance of incorporating structure information into session-level vertex representation learning. Moreover, to evaluate the contribution of associated phrases to utterance-level vertex representation, we remove phrases, denoted as DVAE-GNN w/o phrase. We see that its scores in terms of all the metrics drops sharply, especially the three human evaluation metrics. The reason is that it's difficult to learn high-quality utterance-level vertex representation from a large amount of fine-grained semantic content in open-domain dialogs without any prior information. Moreover, the edge quality of Phrase Graph is worse than DVAE-GNN and DVAE-GNN w/o GNN in ``U-U Appr.''. It indicates DVAE and GNN are effective in capturing utterance-level dialog transitions. The Kappa value is above 0.4, showing moderate agreement among annotators.

\begin{figure*}[htbp]
	\centering
	\includegraphics[width = 0.8\textwidth]{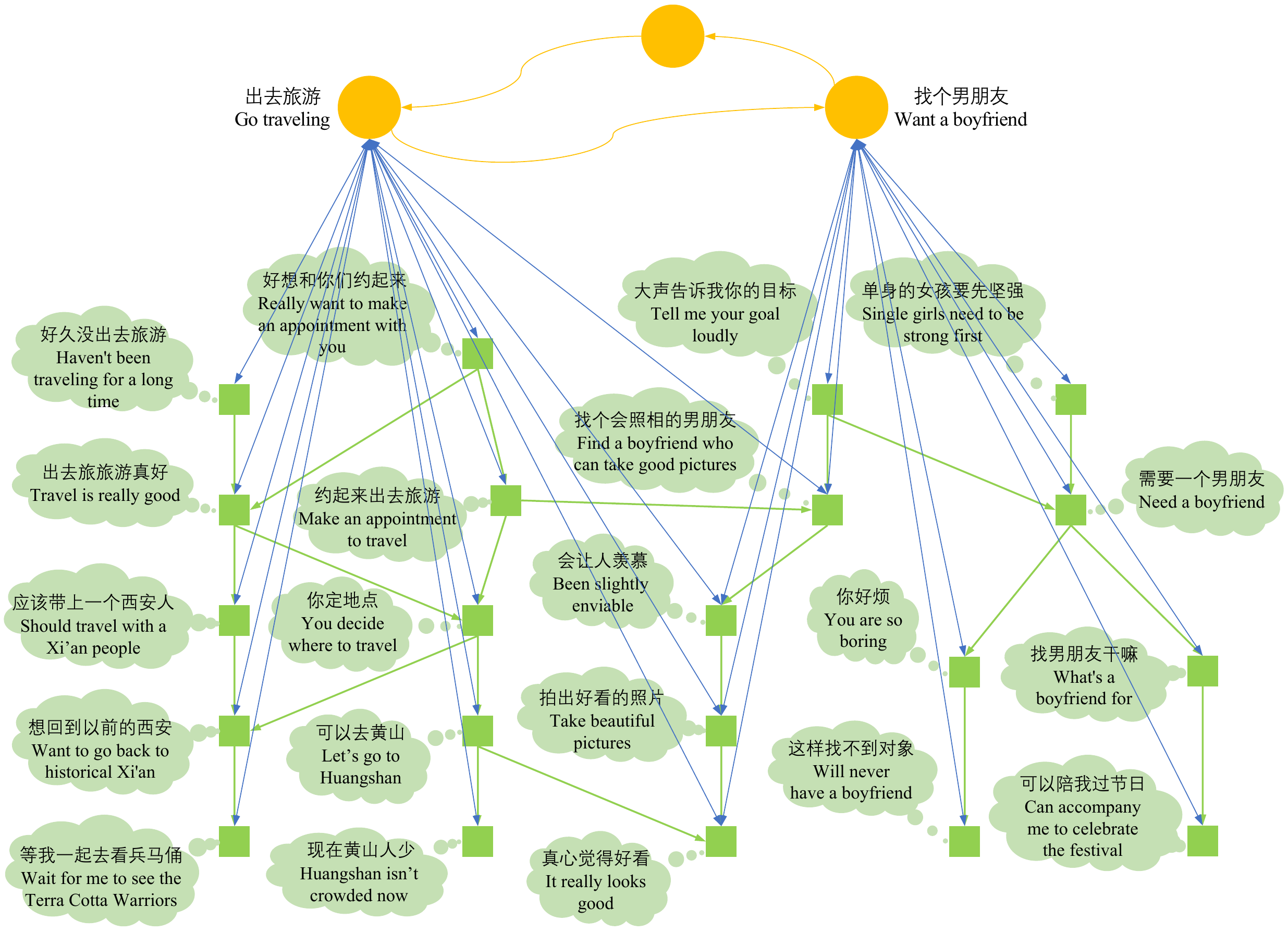}
	\caption{A part of the unified dialog structure graph that extracted from Weibo corpus. Here, we interpret session-level semantics based on their child utterance-level semantic vertices. We translate the original Chinese texts into English language.}
	\label{graphcase1}
\end{figure*}

\subsection{Case Study of Dialog Structure Graph Discovery}
\label{sup:case}
Figure \ref{graphcase1} shows a part of the unified dialog structure graph that discovered from the Weibo corpus. Each yellow-colored circle in this figure represents a session-level semantic vertex with expert interpreted meanings based on the information of top words (from phrases of utterance-level semantic vertices belonging  to this session-level semantic vertex) ranked by TF/IDF. Each green-colored rectangle represents an utterance-level semantic vertex. The directed-arrows between utterance-level semantic vertices represent dialog transitions between states, and the utterance-level semantic vertices within blue dotted-lines are about the same session-level semantic vertex (goal). 

We observe reasonable dialog structures in Figure \ref{graphcase1}. It captures the major interaction logic in dialogs about the goal ``go traveling'', traveling is really good $\rightarrow$ you decide where to travel $\rightarrow$ let's go to Huangshan  $\rightarrow$ comments about Huangshan. Furthermore, it also captures the major logic in dialogs about the goal “want a boyfriend”, need a boyfriend $\rightarrow$ why? $\rightarrow$ he can accompany me to celebrate the festival. Moreover, it captures a dialog goal transition between the goal ``go trveling'' and another goal ``want a boyfriend''.

\section{Experiments for Graph Grounded Dialog Generation}
%We conduct experiments on two publicly available dialog datasets.
%\subsection{Datasets}
To confirm the benefits of discovered dialog structure graph for conversation generation, we use the graph extracted from Weibo corpus to facilitate our system. All the systems (including baselines) are trained on Weibo corpus.

\subsection{Models}
We carefully select the following six baselines.

\noindent \textbf{HRED} It is the hierarchical recurrent encoder-decoder model \cite{Serban2016}, which is commonly used for open-domain dialog generation. 

\noindent \textbf{MMPMS} It is the multi-mapping based neural open-domain conversational model with posterior mapping selection mechanism ~\cite{chen2019generating}, which is a SOTA model on the Weibo Corpus.

%To suit our experiment settings, we translate the commonsense knowledge base into Chinese.
\noindent \textbf{CVAE} It is the Conditional Variational Auto-Encoder based neural open-domain conversational model~\cite{zhao2017learning}, which is a SOTA model for single latent variable based dialog modeling.

\noindent \textbf{VHCR-EI} This variational hierarchical RNN model can learn hierarchical latent variables from open-domain dialogs~\cite{Ghandeharioun2019}. It is a SOTA dialog model with hierarchical VAE.

\noindent \textbf{MemGM} It is the memory-augmented open-domain dialog model~\cite{Tian2019}, which learns to cluster U-R pairs for response generation.

\noindent \textbf{DVRNN-RL} It uses discovered dialog structure graph to facilitate task dialog modeling~\cite{Shi2019}. 

\noindent \textbf{GCS} It is our proposed dialog structure graph grounded dialog system with hierarchical RL. 

\noindent \textbf{GCS w/ UtterG} It is a simpliﬁed version of GCS that just uses the utterance-level graph and utterance-level sub-policy.

We use the same user simulator for RL training of DVRNN-RL, GCS and GCS w/ UtterG. Here, we use the original MMPMS as user simulator because it achieves the best result on the Weibo Corpus. The user simulator is pre-trained on dialog corpus and not updated during policy training. We use the source codes released by the original authors for all the baselines and the simulator. More training details can be found in the Appendix \ref{sup:Details}.

We conduct human evaluation on model-human dialogs. Given a model, we first randomly select an utterance (the first utterance in a session) from test set for the model side to start the conversations with a human turker. Then the human is asked to converse with the selected model till 8 turns (i.e. in total, 16 utterances are generated by the model and the user) are reached. Finally, we obtain 50 model-human dialogs for multi-turn evaluation. Then we randomly sample 200 U-R pairs from the above dialogs for single-turn evaluation.

\subsection{Experiment Settings}
\textbf{Hyper-parameter Setting for Training}
\label{sec:exp_set}
In our experiments, all the models share the same vocabulary (maximum size is 50000 for both Weibo corpus and Douban corpus), initialized word embedding (dimension is 200) with the Tencent AI Lab Embedding. Moreover, bidirectional one-layer GRU-RNN (hidden size is 512) is utilized for all the RNN encoders and RNN decoders. In addition, dropout rate is 0.3, batch size is 32 and optimizer is Adam(lr=2le-3) for all models. During RL training, the discounting weight for rewards is set as 0.95. The MMPMS model for the user simulator employs 10 responding mechanisms. We utilize dependency parse for phrase extraction.\footnote{ai.baidu.com/tech/nlp\_basic/dependency\_parsing} We pre-train the response generator in the Weibo Corpus.
%The three threshold variables $\alpha^{uu}, \alpha^{su}, \alpha^{ss}$ are all set as 0.05.

\begin{table*}[ht] 
	\centering
	\begin{tabular}{l c c  c c  c  c c}
		\toprule
		\multicolumn{1}{l|}{Methods} &  \multicolumn{2}{c|}{\textbf{Coherence}} & \multicolumn{2}{c|}{\textbf{Informativeness}} & \multicolumn{2}{c}{\textbf{Overall Quality}} \\ 
		\cline{2-7} \multicolumn{1}{c|}{}   &  Intra-Cohe.$^*$ &  Appr.$^*$ & \multicolumn{1}{|c}{Info.$^*$} & Dist-1/2$^{\#}$ & \multicolumn{1}{|c}{Enga.$^*$}   & Length$^{\#}$ \\
		\hline
		\multicolumn{1}{l|}{HRED}&  0.54   &\multicolumn{1}{c|}{0.43} & 0.19  &\multicolumn{1}{c|}{0.08/0.26} &0.20 & 5.04  \\
		\multicolumn{1}{l|}{MMPMS}&  0.66   &\multicolumn{1}{c|}{0.45} & 0.50  &\multicolumn{1}{c|}{0.08/0.32} &0.24 & 5.82  \\
		\multicolumn{1}{l|}{CVAE}& 0.58   &\multicolumn{1}{c|}{0.39} & 0.43  &\multicolumn{1}{c|}{0.11/0.38} &0.22 & 7.74  \\
		\multicolumn{1}{l|}{VHCR-EI}&  0.68   &\multicolumn{1}{c|}{0.43} & 0.53  &\multicolumn{1}{c|}{0.12/0.36} &0.28 & 7.30  \\
		\multicolumn{1}{l|}{MemGM}&  0.53   &\multicolumn{1}{c|}{0.37} & 0.34  &\multicolumn{1}{c|}{0.09/0.33} &0.20 & 4.08  \\
		\multicolumn{1}{l|}{DVRNN-RL}& 0.60   &\multicolumn{1}{c|}{0.39} & 0.39  &\multicolumn{1}{c|}{0.06/0.22} & 0.22 & 7.86  \\
		\hline
		\multicolumn{1}{l|}{GCS}& \textbf{1.03} &\multicolumn{1}{c|}{\textbf{0.59}} &\textbf{0.58} & \multicolumn{1}{c|}{\textbf{0.19/0.55}}  &\textbf{0.48} &\textbf{8.00}   \\

		\multicolumn{1}{l|}{GCS w/ UtterG}& 0.93 &\multicolumn{1}{c|}{0.56} & 0.55  &\multicolumn{1}{c|}{0.16/0.47} &0.34 & \textbf{8.00}  \\
		\bottomrule
	\end{tabular}
	\caption{Results for dialogs with humans on Weibo corpus. $*$ or $\#$  denote human or automatic evaluation metrics.}
	\label{tab_dialogweibo}
\end{table*}

\begin{figure}[htbp]
	\centering
	\includegraphics[width = 0.4\textwidth]{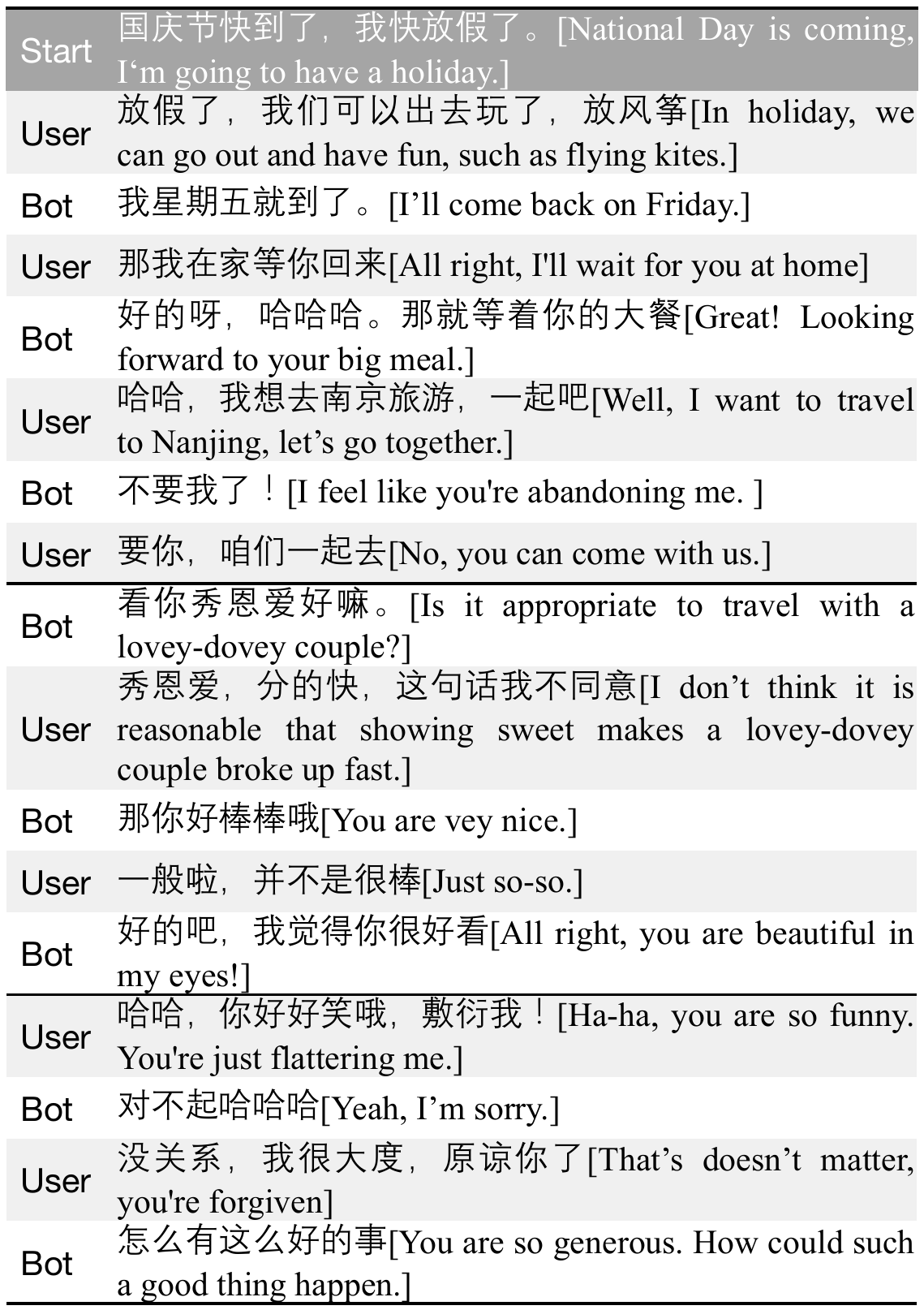}
	\caption{A sample dialog between our dialog system GCS and a human, where“Bot” is our system and “User” is the human. This dialog contains three dialog goals. We translate the original Chinese texts into English language.}
	\label{dialogcase}
\end{figure}

\noindent\textbf{Rewards and Training Procedure for the Graph grounded Conversational System.}
We use the PaddlePaddle framework for the development of our systems.\footnote{paddlepaddle.org.cn/}
We hot-start the response generator by pre-training it before the training of policy module. 
Meanwhile, to make the RL based training process more stable, we employ the A2C method ~\cite{Sutton1998Reinforcement} for model optimization rather than the original policy gradient as done in previous work \cite{li2016deep}. Moreover, during RL training, the parameters of the policy module are updated, and the parameters of response generator and the representation of semantic vertices stay intact.

\subsection{Evaluation Metrics}
Since the proposed system does not aim at predicting the highest-probability response at each turn, but rather the long-term success of a dialog (e.g., coherence), we do not employ BLEU ~\cite{papineni2002bleu} or perplexity for evaluation, and we propose a novel metric (intra-topic coherence) to measure dialog coherence. Finally we use three multi-turn evaluation metrics: intra-topic coherence, dialog engagingness, length of dialog, and three single-turn metrics: appropriateness, informativeness, distinct. For human evaluation, we invite three annotators to conduct evaluation on each case, and we ask them to provide 1/0 (Yes or No) scores for most of the metrics. Moreover, for Intra-Goal Rele., we first ask the annotators to manually segment a dialog by topics (without the use of any tools/models) and then conduct evaluation on each sub-dialog. Notice that system identifiers are masked during human evaluation. We don't evaluate inter-topic coherence as all the models seldom change topics proactively.

\noindent\textbf{Multi-turn Metrics.} 
We use the following metrics: (1)\textbf{Intra-topic Coherence (Intra-Cohe.)} It measures the coherence within a sub-dialog. Common incoherence errors in a sub-dialog include amphora errors across utterances and information inconsistency. ``0'' means that there are more than two incoherence errors in a sub-dialog. ``1'' means that there are one error. ``2'' means that there are no errors. Finally, we compute the average score of all the sub-dialogs. (2)\textbf{Dialog Engagingness (Enga.)} This metric measures how interesting a dialogs is. It is ``1'' if a dialog is interesting and the human is willing to continue the conversation, otherwise ``0''. (3)\textbf{Length of high-quality dialog (Length)} 
%This metric refers to the length of an appropriate dialog. 
A high-quality dialog ends if the model tends to produce dull responses or two consecutive utterances are highly overlapping~\cite{li2016deep}.

\noindent\textbf{Single-turn Metrics.}
We use the following metrics: (1) \textbf{Appropriateness (Appr.)} ``0'' if a response is inappropriate as an reply, otherwise ``1''; (2)\textbf{Informativeness (Info.)}  ``0'' if a response is a ``safe'' response, e.g. ``I don't know'', or it is highly overlapped with context, otherwise ``1''; (3)\textbf{Distinct (Dist.-$i$)} It is an automatic metric for response diversity~\cite{li2015diversity}.

\subsection{Experiment Results}
As shown in Table \ref{tab_dialogweibo}, our proposed GCS significantly outperforms all the baselines in terms of all the metrics except ``Length-of-dialog'' (sign test, p-value $<$ 0.01). It indicates that GCS can generate more coherent, informative and engaging dialogs. Specifically, our system's two sub-policies strategy enables more effective dialog flow control than hierarchical latent variable based VHCR-EI model that performs the best among baselines, as indicated by ``Intra-Cohe'' and ``Enga'' scores. Moreover, our high-quality edges between utterance-level vertices (measured by the metric ``U-U Appr.'' in Table \ref{tab_DSGWeibo}) help GCS to achieve better performance in response appropriateness than DVRNN-RL. In addition, GCS, VHCR-EI, MMPMS and CVAE can obtain better performance in terms of ``Info.'', indicating that latent variable can effectively improve response informativeness. The Kappa value is above 0.4, showing moderate agreement among annotators.

\noindent\textbf{Ablation Study.}
\label{dialog-aba}
In order to evaluate the contribution of session-level vertices, we run GCS with an utterance-level dialog structure graph, denoted as GCS w/ UtterG. Results in Table \ref{tab_dialogweibo} show that its performance in terms of ``Intra-Cohe.'' and ``Enga.'' drops sharply. It demonstrates the necessity of the hierarchical dialog structure graph for dialog coherence and dialog engagingness. Specifically, the removal of the Utter-Goal relevance reward harms the capability of selecting coherent utterance-level vertex sequence.

\subsection{Case Study of Conversation Generation}
\label{case}

Figure \ref{dialogcase} shows a sample dialog between our system “GCS” and a human. We see that our system can generate a coherent, engaging and informative multi-turn dialog. For an in-depth analysis, we manually segment the whole dialog into two sub-dialogs. It can be seen that the first sub-dialog is about “meeting appointment”, and it contains a reasonable dialog logic, I will have a holiday $\rightarrow$ I will arrive $\rightarrow$ wait for you at home $\rightarrow$ look forward to a big meal. And the second sub-dialog is about “joking between friends”, and it also contains a reasonable logic, you are beautiful $\rightarrow$ flattering me $\rightarrow$ I am sorry.

\section{Conclusion}
In this paper, we present an unsupervised model, DVAE-GNN, to discover an unified human-readable hierarchical dialog structure graph from all dialog sessions in a corpus. Specifically, we integrate GNN into VAE to fine-tune utterance-level semantics for more effective recognition of session-level semantic, which is crucial to the success of dialog structure discovery. Further, with the help of the coupling mechanism that binding each utterance-level semantic vertex with a distinct phrase, DVAE-GNN is able to discover a large number of utterance-level semantics. Experiment results demonstrate that DVAE-GNN can discover meaningful dialog structure, and the discovered dialog structure graph can facilitate downstream open-domain dialog systems to generate coherent dialogs.
	
% \begin{acks}
% \end{acks}

\bibliographystyle{ACM-Reference-Format}
\bibliography{emnlp2020}

%\clearpage
\appendix

\section{Datasets and Baselines}
\label{sup:Details}
The links to datasets used in this work are shown as follows. 
\begin{itemize}
    \item Weibo: \url{tcci.ccf.org.cn/conference/2018/dldoc/trainingdata05.zip}
    \item Douban: \url{github.com/baidu/Dialogue/tree/master/DAM}
\end{itemize}
The links to baseline codes used in this work are shown as follows.
\begin{itemize}
    \item HRED: \url{github.com/julianser/hed-dlg-truncated}
    \item MMPMS: \url{github.com/PaddlePaddle/Research/tree/master/NLP/IJCAI2019-MMPMS}
    \item CVAE: \url{github.com/snakeztc/NeuralDialog-CVAE}
    \item VHCR-EI: \url{github.com/natashamjaques/neural_chat}
    \item MemGM: \url{github.com/tianzhiliang/MemoryAugDialog}
    \item DVRNN: \url{github.com/wyshi/Unsupervised-Structure-Learning}
\end{itemize}

\section{Loss Function}
\label{sup:deri}
The proposed DVAE-GNN model consists of two procedures. For a dialog session $X$ that consists of a sequence of $c$ utterances, $X=[x_1, ..., x_c]$, in recognition procedure, we first recognize an utterance-level semantic vertex $z_i$ for each utterance $x_i$, and then recognize a session-level semantic vertex $g$ based on $[z_1, ..., z_c]$. In reconstruction procedure, we regenerate all the utterances in $X$ with the predicted vertices $Z=[z_1, ..., z_c, g]$. Here, we optimize the proposed DVAE-GNN model by maximizing the variational lower-bound:
\begin{equation}
\label{loss}
\nonumber
    \mathbb{E}_{q(Z|X)}[\log p(X|Z)] - KL(q(Z|X) \| p(Z)),
\end{equation}
where $p(Z)$ is the prior uniform distribution of $Z$. 

Specifically, we approximate the first item in the above equation by sampling $Z$ from $q(Z|X)$ and calculate the the negative log-likelihood reconstruction loss. For the second item, we calculate it by: 
\begin{align}
\nonumber
    \sum_{j=1}^{c}KL[q(z_j|x_j) \| p(z_j)] + KL[q(g|z_{1,...,c}) \| p(g)],
\end{align}
where we can calculate each sub-item straightly since $z_{1,...,c}$ and $g$ follow discrete distribution. Below, we provide the derivation of the second item.
\begin{align*}
\notag
	&KL[q(Z|X) \| p(Z)] \\
	&=  \sum_{Z}[\log q(Z|X) - \log p(Z)]q(Z|X) \\
	&=  \sum_{z_{1,...,c}, g}\{ \sum_{j=1}^{c}[\log q(z_j|x_j) - \log p(z_j)] + [\log q(g|z_{1,...,c}) - \\
	&\quad\log p(g)]\}\prod_{i=1}^{c}q(z_i|x_i)q(g|z_{1,...,c})\\
	&=  \sum_{j=1}^{c}\sum_{z_{1,...,c}, g}[\log q(z_j|x_j) - \log p(z_j)]q(z_j|x_j)\prod_{i=1, i\neq j}^{c}q(z_i|x_i)q(g|z_{1,...,c}) \\ 
	&\quad+ \sum_{z_{1,...,c}, g}[\log q(g|z_{1,...,c}) - \log p(g)]q(g|z_{1,...,c})\prod_{i=1}^{c}q(z_i|x_i)\\
	&=  \sum_{j=1}^{c}\sum_{z_j}[\log q(z_j|x_j) - \log p(z_j)]\sum_{z_{[1,...,c]-j}, g}q(z_j|x_j)\prod_{i=1, i\neq j}^{c}q(z_i|x_i)q(g|z_{1,...,c}) \\ 
	&\quad+ \sum_{g}[\log q(g|z_{1,...,c}) - \log p(g)]q(g|z_{1,...,c})\sum_{z_{1,...,c}}\prod_{i=1}^{c}q(z_i|x_i)\\
	&=  \sum_{j=1}^{c}KL[q(z_j|x_j) \| p(z_j)]\sum_{z_{[1,...,c]-j}, g}\prod_{i=1, i\neq j}^{c}q(z_i|x_i)q(g|z_{1,...,c}) \\
	&\quad +  KL[q(g|z_{1,...,c}) \| p(g)]\sum_{z_{1,...,c}}\prod_{i=1}^{c}q(z_i|x_i)\\
	&=  \sum_{j=1}^{c}KL[q(z_j|x_j) \| p(z_j)] + KL[q(g|z_{1,...,c}) \| p(g)]
\end{align*}

\section{A sample dialog structure graph}
\label{sup:case}
\begin{figure*}[htbp]
	\centering
	\includegraphics[width = 0.95\textwidth]{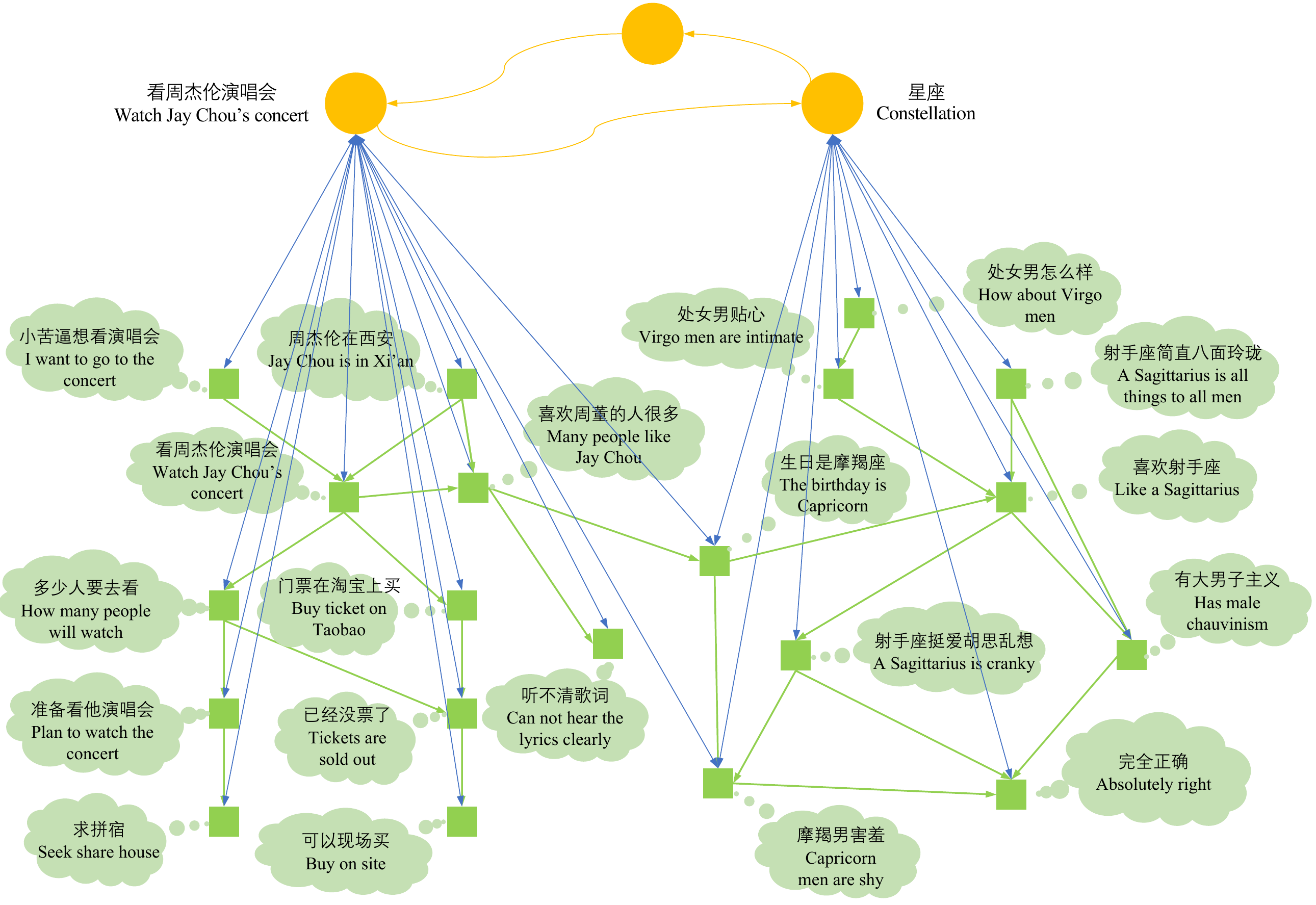}
	\caption{A part of the unified dialog structure graph that extracted from Douban corpus. Here, we interpret session-level semantics based on their child utterance-level semantic vertices. We translate the original Chinese texts into English language.}
	\label{graphcase2}
\end{figure*}

\end{document}